\documentclass{article}

\PassOptionsToPackage{numbers, compress}{natbib}

\usepackage[preprint]{nips}

\usepackage{times}  
\usepackage{helvet}  
\usepackage{courier}  
\usepackage{colortbl} 

\usepackage{graphicx} 
\usepackage{caption} 

\usepackage{amsmath,amsfonts,amssymb}
\usepackage{algorithmicx}

\usepackage{textcomp}
\usepackage{xcolor}
\definecolor{mateGreen}{HTML}{E6F4EA} \definecolor{mateGreenStrong}{HTML}{CFE8D6} \definecolor{mateRed}{HTML}{FDEBEC}
\usepackage{tcolorbox}
\usepackage{tabularx}
\usepackage{float}
\usepackage{longtable}
\usepackage{url} 

\usepackage{booktabs} 
\usepackage{array} 
\usepackage[framemethod=default]{mdframed} 
\usepackage{enumitem}       
\usepackage{listings}

\usepackage{ragged2e} 
\usepackage{makecell} 
\usepackage{algorithm}
\usepackage{algpseudocode} 
\usepackage{multirow} 
\usepackage{amsfonts}
\usepackage{siunitx}
\usepackage{arydshln}
\usepackage{pifont}
\newcommand{\cmark}{\ding{51}}
\newcommand{\xmark}{\ding{55}}

\IfFileExists{fontawesome.sty}{
  \usepackage{fontawesome}
}{
  \newcommand{\faPlusSquare}{\ensuremath{\blacksquare}}
  \newcommand{\faPlusSquareO}{\ensuremath{\square}}
  \newcommand{\faStarO}{\ensuremath{\star}}
  \newcommand{\faMoonO}{\ensuremath{\circ}}
}
\usepackage{wrapfig}   
\usepackage{multicol, dashrule} 
\usepackage{placeins}  
\definecolor{darkgreen}{RGB}{0,128,0}
\definecolor{darkred}{RGB}{180,0,0}

\newcommand\datasetname{\textsc{EthicScaff}}

\title{First, Do No Harm: AI Supervisor Scaffolds Novice Growth in Counselor Education}

\author{%
    \textbf{Chen Xu\textsuperscript{$\heartsuit$ \faPlusSquare}},
    \textbf{Zhenyu Lyu\textsuperscript{$\heartsuit$ \faPlusSquare}}, 
    \textbf{Tian Lan\textsuperscript{$\heartsuit$}}, 
    \textbf{Yang Yi\textsuperscript{$\heartsuit$ \faPlusSquare}},
    \textbf{Yu Ji\textsuperscript{$\heartsuit$ \faPlusSquare}},
    \textbf{Luyao Ji\hspace{0.5mm}\textsuperscript{\rm{\faPlusSquareO}}}
    \textbf{Jian Shen}\textsuperscript{$\heartsuit$ \faPlusSquare}\\
    \textbf{Zhihua Wang}\textsuperscript{$\heartsuit$ \faPlusSquare},
    \textbf{Leyang Cui},\ \ 
    \textbf{Jieshuo Zhang\hspace{0.3mm}\textsuperscript{\faMoonO}},
    \textbf{Xiaohua Wan}, 
    \textbf{Qunxi Dong}\textsuperscript{$\heartsuit$ \faPlusSquare}\\
    \textbf{Minqiang Yang\textsuperscript{\faStarO}}, \ \ 
    \textbf{Juan Wang}\textsuperscript{\faPlusSquareO}, \ \ 
    \textbf{Xiuling Liu}\textsuperscript{\faMoonO}, \ \ 
    \textbf{Bin Hu}\textsuperscript{$\heartsuit$ \faPlusSquare \ \faStarO} \\
    \textsuperscript{$\heartsuit$} Key Laboratory of Brain Health Intelligent Evaluation and Intervention, \\ Ministry of Education  (Beijing Institute of Technology) \\
    \textsuperscript{\faPlusSquare} School of Medical Science and Engineering, Beijing Institute of Technology \\
    \textsuperscript{\faPlusSquareO} Chinese People’s Liberation Army General Hospital \\
    \textsuperscript{\faStarO} Lanzhou University \textsuperscript{\faMoonO} Hebei University \\
}

\begin{document}

\maketitle

\begin{abstract}
The most dangerous mistakes a novice counselor makes are not the obvious ones: they are utterances that sound caring while quietly violating professional ethics and leaving vulnerable clients less protected. 
%
We build an AI supervisor that does not replace novice counselors, but grows them—teaching them to internalize ethical violations they would otherwise never notice. What makes this supervisor non-trivial is not detection but teaching: it must locate the ethical-violating utterance, diagnose the ethical violation against APA principles, and deliver feedback that explains not just what went wrong, but why it is risky and how to respond differently. 
%
The core obstacle is that (1) ethical violations are by nature unlabeled in real clinical data, and (2) existing AI counselors trained only to match correct answers will never learn to teach. We resolve both at once: a controllable AI novice that intentionally enacts predefined mistake categories makes supervision labels a natural byproduct of generation, yielding \datasetname{}, a 9,915-instance human-in-the-loop dataset; and GRPO under a Novice Growth Reward (NGR) optimizes the supervisor not for answer correctness but for whether a weaker novice model actually improves after reading its explanation. 
%
Experiments show that a novice guided by our supervisor outperforms an unguided peer on clinical metrics, and that teaching-oriented optimization via NGR further sharpens the supervisor's own ethical detection. In a user study with novice counseling-psychology students, participants show significant self-efficacy gains across all eight assessed competencies after receiving AI supervisory feedback, demonstrating that the scaffold transfers from simulation to real-world practice.
\end{abstract}


\section{Introduction}

\paragraph{AI for Novice Counselor Education}
The field of mental health faces a persistent imbalance between growing demand and the limited availability of trained counselors~\cite{WHO2023mentalhealth}. Although large language models (LLMs) have been explored as low-cost providers of mental health support~\cite{xiao2024healme, na2024cbt, xie2025psydt, wang2025emotional,shen2024large}, deploying them directly in patient-facing treatment remains ethically and clinically fragile. Patient-facing LLMs raise concerns about professional responsibility, safety, privacy, and therapeutic alliance~\cite{ong2024ethical,raile2024usefulness,hager2024evaluation,chung2023challenges,nichol2023not,choudhury2024large,scholich2025comparison}. 
A safer and more scalable role for AI is therefore to accelerate counselor education---the systematic training process that prepares novices to master therapeutic techniques and treatment strategies before working with real clients~\cite{wang2024patient,akkurt2025learning,beeson2025pilot, wang2025annaagent,li2024leveraging,brugge2024large}.

\textbf{``First, Do No Harm''} Prior work expands practice in therapeutic techniques but often underemphasizes ethics, the field's most fundamental obligation. This foundational principle~\cite{smith2005origin, sokol2013first} underscores why ethics education must take priority in counselor training. Well-intentioned novices can cause harm they cannot yet recognize~\cite{psychologists2010ethical,vybiral2024negative,gerke2020frequencies}. Consider a novice counselor who, hearing a client from a collectivist culture express guilt about prioritizing family over personal goals, responds: ``You need to put yourself first. That's healthy boundaries.'' The advice sounds empowering, yet it imposes individualistic values, invalidates the client's cultural identity, and may deepen their distress. Similarly, when a client hints at passive suicidal ideation, saying ``Sometimes I wonder if people would be better off without me,'' a novice might respond: ``You don't need to think that way. I'm sure your family loves you.'' This sounds caring, but it is premature reassurance that silences disclosure and bypasses critical risk assessment. The danger is not that novices intend harm, but that they cannot yet recognize it.

\begin{wrapfigure}{r}{0.5\textwidth}
  \vspace{-15pt}
  \centering
  \includegraphics[width=\linewidth]{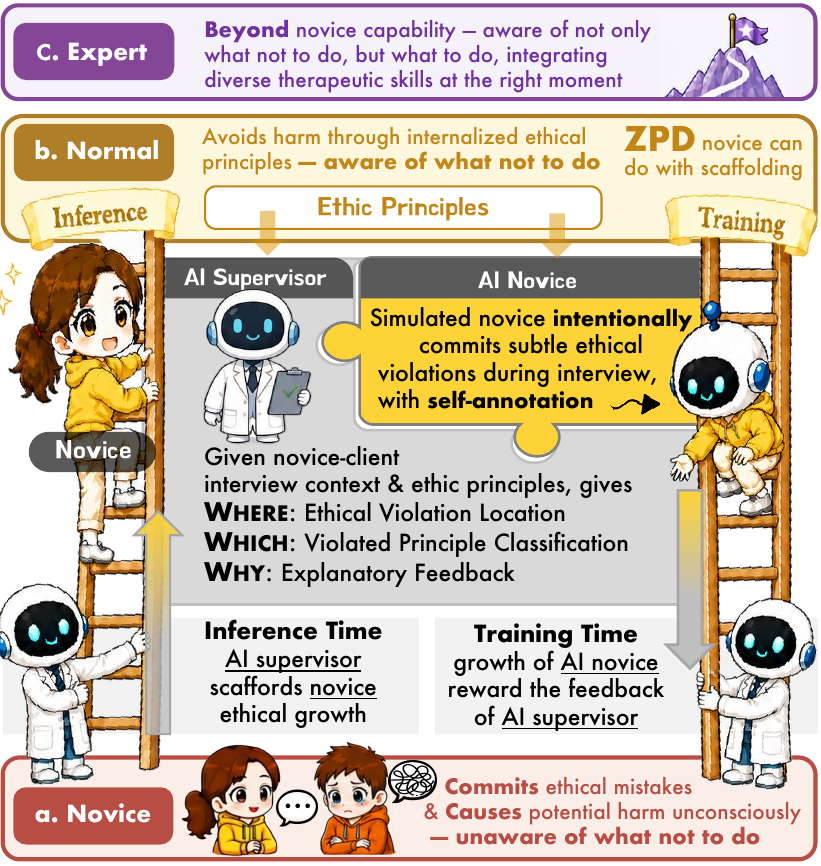}
  \caption{Illustration of how our AI supervisor supports ethical learning within the zone of proximal development (ZPD) and helps novices grow into normal ethical practice. A novice counselor may initially make ethical mistakes without recognizing them, but with guidance can gradually improve. During training, a simulated AI novice intentionally makes mistakes and learns from the supervisor's ``scaffold'' feedback; during inference, the trained supervisor reviews a counseling dialogue and provides structured feedback on where the ethical-violating occurs, what mistake it reflects, why it violates ethical principles, and how to avoid it.}
  \vspace{-0.07em}
  \label{fig:intro}
  \vspace{-25pt}
\end{wrapfigure}

\paragraph{AI Supervision Fits as Ethical Learning Scaffold}
How do novices learn to internalize ethical principles? For novice, learning what not to do is more achievable than mastering what to do. Expert-level judgment (Fig.\ref{fig:intro}c), such as knowing the best therapeutic response, requires weighing countless contextual factors and varies by approach and counselor style, resisting any fixed standard~\cite{leichsenring2017cognitive,kramer2020individualizing,leichsenring2018gold}. Ethical constraints, by contrast, are fundamentally more tractable (Fig.\ref{fig:intro}b): the American Psychological Association (APA) provides a finite set of universal principles that apply across all approaches, and violations follow recognizable patterns~\cite{psychologists2010ethical,vybiral2024negative,gerke2020frequencies}. However, a difficulty remains in practice. Traditional supervision relies on repeated exposure under expert guidance, but is slow, resource-intensive, and inconsistent. The core challenge is what Vygotsky termed the Zone of Proximal Development (ZPD)~\cite{vygotsky1978mind,wood1976role}: novices cannot see their violations alone yet can learn to recognize them with guidance (Fig.\ref{fig:intro}a $\rightarrow$ Fig.\ref{fig:intro}b) . Therefore, this work proposes an ethical learning framework where simulated patients enable repeated exposure to novices' own mistakes, while an AI supervisor surfaces hidden violations and delivers feedback clarifying where violations occur, which principles are breached, and why.

\paragraph{Two Obstacles for Training AI Supervisors} 
\textit{First, there is no suitable training data.} Existing open-source counseling datasets~\cite{lee2024cactus,liu2023chatcounselor} were curated to train high-quality counselors, not to teach novices what not to do. Their dialogues are expert-validated, ethically vetted, and filtered to exclude the mistakes we need novices to recognize. Safety-focused datasets~\cite{ji2024pkusaferlhf,ganguli2022redteaming}, while containing harmful content, target overt violations (e.g., abuse, discrimination) rather than the nuanced ethical missteps that counseling novices typically make. More critically, even if subtle violations were present, these datasets contain no ground-truth annotations for supervision. There are no labels indicating where violations occur or which ethical principles are compromised. Most importantly, there are no explanations of why the identified where and which constitute violations, or how to respond differently. Such explanatory scaffolding is essential for human learners to internalize ethical principles, not merely for models to eliminate risks and respond safely to replace human counselors~\cite{chung2023challenges,scholich2025comparison,raile2024usefulness}. The supervision task demands both dialogues containing subtle ethics violations and rich pedagogical annotations. Yet existing datasets provide neither.

\textit{Second, there is a mismatch of training objectives.} Even if suitable data existed, the standard training objective is fundamentally misaligned with the goal of supervision. Existing datasets optimize AI counselors to generate correct responses, evaluated by matching against gold-standard answers~\cite{papineni-etal-2002-bleu,lee2024cactus,liu2023chatcounselor}---these works well for training AI counselors, but not AI supervisors. A supervisor's role is not to provide correct answers, but to help novices internalize principles so they can recognize and avoid violations in future cases, rather than merely memorizing corrections for cases they have already encountered. For instance, directly telling a novice ``line 3 violates the non-judgment principle'' is perfectly ``correct'' feedback, yet it fails to explain why this principle matters or how violations harm clients. A pedagogically effective supervisor, returning to the earlier case, would instead explain: ``When you said `You don't need to think that way. I'm sure your family loves you,' this premature reassurance violates non-judgment because it dismisses the client's lived experience and closes the space for risk assessment. The client may interpret your response as `my feelings aren't valid here' and stop sharing suicidal thoughts.'' This is the distinction between teaching to the test and teaching to think~\cite{jensen2014teaching, zakharov2021teaching}.

\paragraph{Our Approach and Contributions}

Motivated by all above, we (1) propose, to our knowledge, the first learning framework dedicated to ethical supervision in novice counselor education. The framework is organized around a tripartite structure: a simulated patient creates realistic counseling contexts, a trainee counselor produces novice-level responses, and an AI supervisor provides scaffolded ethical guidance. Rather than simply judging whether a response is correct, the supervisor teaches the novice by identifying where the ethical-violating utterance occurs, which ethical mistake it reflects, and why it is risky through explanatory pedagogical feedback grounded in professional ethics, as shown in Figure~\ref{fig:intro}.

To train such AI supervisors, for the lack of suitable ethics-centered supervision data, we (2) construct \datasetname{}, an dialogue-feedback dataset driven by ethics-violating novice responses. The dataset enables training supervisors to locate ethical-violating utterances, classify novice-counselor mistakes, and provide explanatory feedback for ethical growth. For the misalignment between standard answer-matching objectives and novice growth-oriented supervision, we (3) introduce a two-stage supervisor training framework that combines SFT with Novice Growth-Guided RL, encouraging feedback that is not only correct but also genuinely educational for weaker novice counselors.

Empirically, we (4) demonstrate that better teachers produce better students across four dimensions. First, better supervision leads to better counseling behavior: feedback from the trained supervisor helps novice counselors produce responses that are more collaborative, clinically appropriate, empathic, and alliance-building. Second, supervision-oriented training sharpens ethical judgment: models trained on \datasetname{} become substantially better at recognizing both where harmful counselor utterances occur and which ethical mistakes they reflect. Third, the dataset and training framework provide a meaningful supervision signal: ablations show that targeted supervision data, learner-centered optimization, and data-quality refinement each contribute to stronger and more reliable supervisory competence. Fourth, these gains extend to human-facing training: automatic evaluation, expert judgment, and novice self-efficacy results all indicate that the supervisor provides higher-quality feedback and can serve as a practical scaffold for counselor learning.

\section{Related Works}
\paragraph{LLM Critique and Supervisory Feedback}
Recent work has increasingly studied LLMs as critics rather than only generators. This includes benchmarks and training methods for critique quality~\cite{lin2024criticbench, lan2024criticeval,lan2024training,xu2026bridging, zhang2025codecriticbench}, evaluator-specialized models such as Prometheus~\cite{kim2023prometheus, kim2024prometheus}, and reinforcement-learning approaches that explicitly optimize critique behavior~\cite{xi2025critique}. In parallel, feedback generation has also been explored in counselor-training settings, where LLM-generated feedback links strategy use to evaluative comments~\cite{chaszczewicz2024multi}. Our work is related to this line but differs in both target and granularity. We study psychotherapy supervision centered on ethical violations and high-risk novice mistakes, where the supervisor must not only judge response quality, but also identify problematic utterances, classify mistake types, and give clinically grounded, teachable feedback.

\paragraph{Teacher-Student Learning from Feedback}
Another related line studies teacher-student frameworks in which feedback quality is judged by its effect on a learner. In LLM alignment, a stronger or privileged teacher can provide supervision that allows a weaker student to improve through imitation or iterative fine-tuning~\cite{choudhurybetter}. In education-oriented settings, simulated students have been used to train instructors~\cite{markel2023gpteach}, tutor LLMs have been optimized to maximize predicted student learning outcomes across dialogues~\cite{scarlatos2025training}, and automated feedback systems have been shown to improve teachers' uptake of student ideas at scale~\cite{demszky2024automated}. Work on novice peer counselors likewise explores multi-level feedback generation under high-stakes constraints~\cite{chaszczewicz2024multi}. Our setting is different: we treat novice improvement as the signal for optimizing the supervisor, so the goal is not only correctness but educational effectiveness for weaker learners.


\section{Methodology}
\label{sec:methodology}

\begin{figure*}[!t]
\centering
\includegraphics[width=\textwidth]{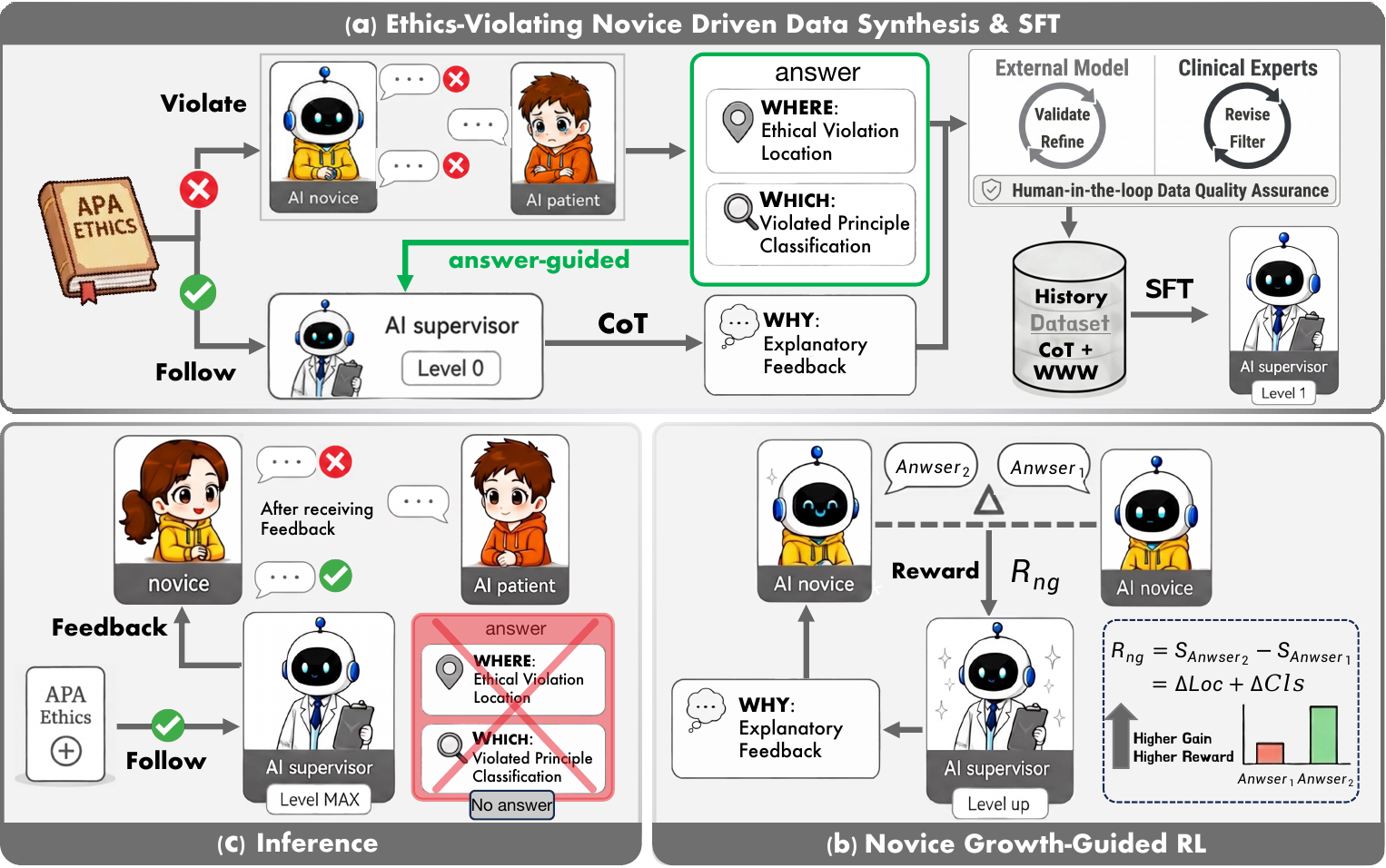}
\caption{Overview of the proposed bidirectional supervision scaffold. (a) Ethics-violating novice data synthesis builds \datasetname{} from controllable novice-patient dialogues: a Level-0 supervisor generates answer-guided supervision over where the ethical-violating utterance occurs, which violated principle category it reflects, and why it is problematic; external-model validation and clinical-expert review refine the data, which are then used for SFT to initialize a Level-1 supervisor. (b) Novice Growth-Guided RL further trains the supervisor with GRPO, rewarding explanations that improve a weaker novice's localization and classification performance without directly disclosing the answer. (c) At inference time, the trained supervisor reviews a counseling dialogue and provides scaffolded feedback to support the novice counselor's ethical reflection and self-correction.}
\label{fig:method}
\end{figure*}

\paragraph{Task Definition}
Psychological counseling supervision is not merely error detection, but a mechanism for ethical growth. We therefore formulate an LLM-based psychological counseling supervision mechanism that observes counselor-client dialogues and provides professional feedback for novice counselors.
Let $\mathcal{H} = \{u_1, u_2, ..., u_n\}$ denote the dialogue history, where $u_i$ represents the utterance in the $i$-th turn. Let $\mathcal{S}$ denote the set of mistaken utterances, $\mathcal{M}$ the set of mistake categories, and $\mathcal{F}$ the set of feedback.
We formalize the supervision task as the following joint probability distribution $P(f, m, s | \mathcal{H}) ,$ where $f \in \mathcal{F}$, $m \in \mathcal{M}$, and $s \in \mathcal{S}$.
According to the chain rule of probability, this joint distribution can be decomposed into three sequential conditional probabilities. First, the model performs Ethical Violation Location by computing $P(s | \mathcal{H})$, which locates the specific utterance containing the counselor's mistake based on the dialogue history. Second, given the identified mistaken utterance, the model conducts Violated Principle Classification through $P(m | \mathcal{H}, s)$, determining the clinical category of the identified problem. Finally, the model generates explanatory pedagogical feedback via Feedback Synthesis, modeled as $P(f | \mathcal{H}, s, m)$, which synthesizes all available information to produce actionable guidance.
Therefore, the complete generation process can be expressed as:
\begin{equation}
P(f, m, s | \mathcal{H}) = P(s | \mathcal{H}) \cdot P(m | \mathcal{H}, s) \cdot P(f | \mathcal{H}, s, m)
\end{equation}
This decomposition mirrors the three core steps of real clinical supervision: locating the problematic utterance, identifying the underlying violation, and delivering constructive feedback.

\subsection{Ethics-Violating Novice Driven Data Synthesis}

Given the scarcity of high-quality supervision data in psychological counseling, together with privacy constraints, sample imbalance in real clinical data, and the fact that ethical violations are often subtle and rarely labeled, we need a controllable way to make such failures observable and teachable. We therefore collaborate with clinical experts, who distilled 15 common novice-counselor violation types from the \emph{APA Principles}~\cite{psychologists2010ethical}, and add one ``Exemplary Practice'' category to construct Counselor Ethical Principles $\mathbb{P} = \{P_1, P_2, ..., P_{16}\}$, where each principle $P_k$ codifies a novice-relevant failure mode together with its corrective supervision logic as a triple:
$
P_k = (cat_k, beh_k, strat_k)
$
where $cat_k$ denotes the category name, $beh_k$ describes the violation behavior, and $strat_k$ specifies the corresponding supervision explanatory strategy; together, these principles define both what should be corrected and what ethically grounded practice looks like. Table~\ref{tab:maladaptive_example} shows one representative example.
We then use these principles to build an ethical-violation-driven data synthesis framework that generates controlled dialogue-feedback pairs in which hidden ethical failures become visible, teachable, and learnable.

\begin{wraptable}{r}{0.5\textwidth}
  \vspace{-12pt}
  \caption{Example of Counselor Ethical Principles}
  \label{tab:maladaptive_example}
  \centering
  \footnotesize
  \renewcommand{\arraystretch}{1.1}
  \setlength{\tabcolsep}{2pt}
\begin{tabular}{p{0.2\linewidth} p{0.8\linewidth}}
\toprule
\textbf{Item} & \textbf{Content} \\
\midrule
Category & Cultural Insensitivity \\
Ethics-violating Behavior & The counselor overlooks how the client's culture, gender, religion, or other identities shape their distress, leading to biased understanding or inappropriate intervention. \\
Explanatory  Strategy & Encourage cultural humility: ask how cultural factors shape the client's worldview and distress, avoid assumptions, and treat respect for diversity as a core professional value. \\
\bottomrule
\end{tabular}
\vspace{-8pt}
\end{wraptable}

\subsubsection{Role-Conditioned Behavior Simulation}
The framework is driven by two independent sets: the Counselor Ethical Principles $\mathbb{P}$ and the Client Case Set~\cite{wang2024patient} $\mathbb{C} = \{C_1, C_2, ..., C_{106}\}$. Together, they let us ground supervision in professional ethical constraints while preserving realistic variation in client context.
For the ethics-violating novice counselor $T_k$, we instantiate a specialized counselor simulator powered by carefully designed prompts for each $P_k \in \mathbb{P}$. This simulator is prompted to consistently exhibit the $k$-th specific violation during conversations, creating the kind of novice error that a supervisor must learn to catch and teach through. Its behavior in the $i$-th dialogue turn is modeled as: $u_i^T = T_k(\mathcal{H}_{i-1}, cat_k, beh_k),$ where $u_i^T$ is the counselor's utterance in the $i$-th turn, $\mathcal{H}_{i-1} = \{u_1, ..., u_{i-1}\}$ represents the dialogue history up to turn $i-1$, and $cat_k$ and $beh_k$ extracted from $P_k$ determine the simulator's behavior pattern.
Correspondingly, for each client case $C_j \in \mathbb{C}$, we instantiate a behavior-sensitive simulated client $C_j$. This simulator possesses unique background and personality traits while being capable of producing contextually appropriate responses to the counselor's specific mistakes. This interaction is crucial because many ethical violations only become legible when their downstream effect on the client is placed in context. Its behavior is modeled as: $u_i^C = C_j(\mathcal{H}_{i-1} \oplus u_i^T, Case_j), $where $u_i^C$ is the client's response, $\oplus$ denotes sequence concatenation, $Case_j$ contains the client's information.

\subsubsection{Dialogue Generation and Feedback Synthesis}

In the dialogue generation stage, the system orchestrates multi-turn interactions between an ethics-violating novice counselor $T_k$ and a simulated client $C_j$, producing a complete dialogue history:
\begin{equation}
D_{j,k} = G_{dialogue}(T_k, C_j) = \{(u_1^T, u_1^C), ..., (u_n^T, u_n^C)\}
\end{equation}
Following dialogue generation, a LLM-based clinical supervisor $\mathcal{S}$ with an ``omniscient perspective'' reviews the entire dialogue process $D_{j,k}$ and generates structured critical feedback based on predefined violation definition $P_k$:
\begin{equation}
F_{j,k} = \mathcal{S}(D_{j,k}, P_k) = (s_{j,k}, m_{j,k}, f_{j,k})
\end{equation}
where the feedback $F_{j,k}$ comprises three components: $s_{j,k}$ identifies the problematic sentence by precisely quoting the specific utterance embodying the violation; $m_{j,k}$ denotes the violation category, explicitly stating the category of error committed by the novice counselor; and $f_{j,k}$ provides concrete and actionable feedback such as response strategies and improvement suggestions. In this sense, the synthesized target is not merely a label, but a compact instance of supervision: what went wrong, why it matters, and how the novice can improve.
Finally, the dialogue and feedback are combined into a complete data pair $P_{j,k} = (D_{j,k}, F_{j,k})$, which collectively forms the dataset required for training model. Specifically, we adopt a full cross-combination strategy over 106 client types and 16 counselor-behavior patterns, and sample each combination 6 times with DeepSeek-V3, yielding 10,176 automatically annotated instances with both violation category labels and problematic utterance labels. This synthesis process therefore supplies not only scale, but also the pedagogically structured supervision signal needed to train an AI supervisor for novice growth.

\subsection{Data Quality Assurance}

To ensure data quality, we process the initial 10,176 records with systematic cleaning, Validator-Guided Refinement (VGR), and expert manual review, yielding 9,915 final samples. The goal here is not only label cleanliness, but educational fidelity: the feedback should model the kind of progressive, ethical, and actionable supervision that can actually help a novice improve. In VGR, GPT-4o acts as both refiner and validator in a closed loop: it first improves the feedback $f_{j,k}$ into $f'_{j,k}$ given the dialogue $D_{j,k}$ and structured label $(s_{j,k}, m_{j,k})$, then audits the refined feedback against an expert-designed rubric. Two clinical psychology experts define checklist criteria across four supervision dimensions: progressivity, actionability, ethicality, and supportiveness. A sample is accepted only if all criteria are satisfied; otherwise, refinement is retried up to $N_{\text{retry}}$ times. Samples that still fail are manually refined by two licensed clinical psychology experts through independent review and consensus revision. Inter-rater agreement between automated and manual refinement reaches Cohen's kappa $\kappa = 0.87$, indicating substantial agreement and supporting the reliability of VGR.

\subsection{Supervisor Model Training}
We train the supervisor in two stages. We first perform SFT on \datasetname{} so that the model learns the supervision format and acquires the basic ability to localize ethical-violating utterances, classify mistake types, and provide targeted feedback. We then apply RLVR with GRPO, where the key objective is not only to produce correct supervision outputs, but also to generate explanations that are genuinely \emph{teachable}. This stage completes the bidirectional scaffold: the same system that guides novices at inference time is optimized during training for whether its feedback can actually produce novice growth.

\textbf{Novice Growth-Guided RL.}
Our central signal is the Novice Growth Reward(NGR). Given the supervisor's explanation, we let a frozen novice model (Llama3.1-8B) solve the ethical violation localization and classification task before and after reading that explanation. The supervisor is rewarded when its explanation improves the novice's performance without directly revealing the answer. In this way, we optimize for feedback that is instructional and transferable, rather than merely answer-matching. The overall reward is:
\begin{equation}
R_{\text{total}} = w_1 \cdot R_{\text{format}} + w_2 \cdot R_{\text{loc}} + w_3 \cdot R_{\text{classify}} + w_4 \cdot R_{\text{ng}} .
\end{equation}
Let $(\hat{s}^{\,b}, \hat{m}^{\,b})$ and $(\hat{s}^{\,a}, \hat{m}^{\,a})$ denote the novice predictions before and after reading the explanation. We define the Novice Growth Reward as
\begin{equation}
R_{\text{ng}} = \lambda_1 \cdot \big[\mathrm{Prec}(\hat{s}^{\,a}) - \mathrm{Prec}(\hat{s}^{\,b})\big] + \lambda_2 \cdot \big[\mathrm{Acc}(\hat{m}^{\,a}) - \mathrm{Acc}(\hat{m}^{\,b})\big].
\end{equation}
This design encourages the supervisor to provide stepwise, non-leaking guidance that helps a weaker learner improve.

\section{Experiments \& Analysis}
\label{sec:experiments}
We evaluate the proposed supervision framework along four axes: downstream counselor behavior, ethical judgment on ethical violation classification and utterance localization, component contributions through ablations, and human-facing training value through automatic, expert, and novice self-efficacy evaluations.

\subsection{Better Teachers Produce Better Students}
We first explore whether stronger supervision actually produces stronger downstream counselors. Following the Counselor-Level Evaluation protocol in PsycEval~\cite{pan2026psycheval}, we compare four counselor conditions. Besides the novice counselor, we include a stronger unguided reference, denoted as the normal counselor, instantiated by PsyDT~\cite{xie2025psydt}. Table~\ref{tab:comparison} shows that adding supervision already improves a novice counselor across all six psychological metrics. Using the vanilla Qwen-3-8B model as the baseline supervisor yields consistent gains, while our \datasetname{}-tuned Qwen-3-8B produces further improvements on every metric.
The largest gains appear on MITI, PSC, and WAI, suggesting that better supervision does not merely make responses sound nicer, but helps novices become more collaborative, clinically appropriate, and alliance-building. Notably, our supervised novice approaches the normal counselor on most dimensions and slightly surpasses it on HTAIS and WAI. This supports our central claim: a better teacher can indeed bring out a better student.

\begin{table}[h]
\vspace{-8pt}
\centering
\small
\caption{Performance Comparison Across Counseling Quality Metrics Under Different Counselor Conditions and Feedback Settings. All metrics are higher-is-better.}
\label{tab:comparison}
\setlength{\tabcolsep}{3.5pt}
\renewcommand{\arraystretch}{1.05}
\begin{tabular}{>{\raggedright\arraybackslash}p{2.15cm}>{\raggedright\arraybackslash}p{3.8cm}cccccc}
\toprule
\textbf{Condition} & \textbf{Method} & \textbf{EFT-TFS} & \textbf{HTAIS} & \textbf{MITI} & \textbf{PSC} & \textbf{TES} & \textbf{WAI} \\
\midrule
\multirow{2}{=}{\centering\textbf{No\\Feedback}} & Novice Counselor     & 2.21 & 4.79 & 5.62 & 5.98 & 7.65 & 4.68 \\
 & Normal Counselor (PsyDT)     & \textbf{3.01} & 5.34 & \textbf{7.35} & \textbf{7.14} & \textbf{8.73} & 6.00 \\
 \hdashline[2pt/2pt]

\multirow{2}{=}{\centering\textbf{With\\Feedback}} & Baseline Feedback   & 2.78 & 5.21 & 6.16 & 6.31 & 7.80 & 5.05 \\
 & \textbf{Our Feedback}        & 2.97 & \textbf{5.35} & 7.22 & 7.01 & 8.54 & \textbf{6.06} \\
\midrule

\multirow{2}{=}{\centering\textbf{Improvement\\over Novice}} & Baseline Feedback    & +0.57 & +0.42 & +0.54 & +0.33 & +0.15 & +0.37 \\
 & \textbf{Our Feedback}        & \textbf{+0.76} & \textbf{+0.56} & \textbf{+1.60} & \textbf{+1.03} & \textbf{+0.89} & \textbf{+1.38} \\
\bottomrule
\end{tabular}

\end{table}
\vspace{-5mm}

\subsection{Supervision Training Sharpens Ethical Judgment}

We next explore whether supervision-oriented training can sharpen the model's ethical judgment: locating problematic counselor utterances and diagnosing the underlying mistake category. Table~\ref{tab:mistake_classification} reveals a clear asymmetry between the two objective tasks. For untuned open-source models, violated principle classification remains substantially weaker than proprietary baselines, while ethical violation location is relatively more competitive. This suggests that localization relies more on general discourse understanding, whereas fine-grained category discrimination requires stronger task-specific supervision.
After incorporating \datasetname{}, open-source models improve dramatically on both tasks and overtake the proprietary baselines by a large margin. On Task~1, Qwen3-14B (w/ \datasetname{}, Full) achieves 94.37\% F1-score, far above GPT-4o with RAG (73.12\%) and DeepSeek-V3 (46.85\%). On Task~2, Qwen3-8B (w/ \datasetname{}, Full) reaches the best overall F1-score of 74.24\%, together with the strongest Jaccard (63.03\%) and EMR (18.40\%), while Qwen3-14B (w/ \datasetname{}, Full) remains highly competitive at 73.25\% F1-score.

These gains suggest that the proposed supervision-training framework equips models with ethical judgment that is largely absent from general-purpose prompting or retrieval. The Full variants further improve over SFT-only variants, indicating that Novice Growth Reward not only optimizes feedback for learners but also strengthens the supervisor's own diagnostic competence. At the same time, localization still exhibits a high-recall bias: several RAG-based variants obtain the highest recall (up to 97.15\%) but do not yield the best F1-score or EMR, indicating a tendency to over-predict problematic sentences. In contrast, models trained on \datasetname{} achieve a better trade-off.

\begin{table}[ht]
  \caption{Performance comparison on objective evaluation tasks. For each model family (Llama3.1-8B, Qwen3-8B, and Qwen3-14B), the best result within the group is highlighted in \textbf{bold}.}
  \vspace{5pt}
  \centering
  
  \resizebox{\textwidth}{!}{%
  \renewcommand{\arraystretch}{1.0}
  \begin{tabular}{
  p{3cm} p{3.4cm}|
  c c c c|
  c c c c c}
  \hline
  \multirow{2}{*}{\textbf{Model Family}} & \multirow{2}{*}{\textbf{Variant}} &
  \multicolumn{4}{c|}{\textbf{Task 1: Violated Principle Classification}} &
  \multicolumn{5}{c}{\textbf{Task 2: Ethical Violation Location}} \\[0.3ex]
  \cline{3-6}\cline{7-11}
  & & {Accuracy} & {Precision} & {Recall} & {F1-Score} &
  {Precision} & {Recall} & {F1-Score} & {Jaccard} & {EMR} \\[0.3ex]
  \hline
  
  \rowcolor{gray!12}
  \multicolumn{11}{l}{\textit{Closed-Source Models}} \\[0.2ex]
  \multirow{2}{*}{GPT-4o}
  & Base & 51.18 & 62.69 & 51.18 & 45.66 & 62.19 & 89.93 & 69.48 & 57.63 & 13.68 \\[0.3ex]
  & RAG & 73.61 & 81.52 & 73.61 & 73.12 & 61.90 & 95.17 & 71.76 & 59.59 & 14.10 \\[0.3ex]
  Claude-Sonnet-4
  & Base & 54.59 & 65.82 & 54.59 & 47.44 & 59.57 & 94.67 & 70.04 & 58.04 & 13.14 \\[0.3ex]
  
  \hline
  \rowcolor{gray!12}
  \multicolumn{11}{l}{\textit{Open-Source Models}} \\[0.2ex]
  DeepSeek-V3-685B
  & Base & 52.14 & 64.04 & 52.14 & 46.85 & 60.71 & 97.05 & 71.85 & 59.66 & 13.57 \\[0.3ex]
  
  \hdashline[1pt/2pt]
  \multirow{4}{*}{Llama3.1-8B}
  & Base & 29.39 & 43.73 & 29.39 & 25.96 & 59.40 & 94.31 & 69.91 & 57.28 & 10.01 \\[0.3ex]
  & Few-shot & 18.12 & 22.00 & 18.12 & 16.07 & 59.75 & 94.92 & 70.49 & 57.86 & 10.57 \\[0.3ex]
  & RAG & 57.22 & 62.27 & 57.22 & 54.98 & 59.95 & \textbf{96.49} & 71.25 & 58.81 & 12.30 \\[0.3ex]
  \rowcolor{mateGreen}
  & w/ {\small\datasetname{}} & \textbf{67.96} & \textbf{63.22} & \textbf{67.96} & \textbf{62.22} & \textbf{63.64} & 93.55 & \textbf{72.70} & \textbf{61.19} & \textbf{17.13} \\[0.3ex]
  
  \hdashline[1pt/2pt]
  \multirow{6}{*}{Qwen3-8B}
  & Base & 23.33 & 44.16 & 23.33 & 22.18 & 54.78 & 60.76 & 53.56 & 40.52 & 3.87 \\[0.3ex]
  & Few-shot & 33.55 & 46.13 & 33.55 & 31.58 & 56.25 & \textbf{95.64} & 68.00 & 54.98 & 9.94 \\[0.3ex]
  & CoT & 50.22 & 57.03 & 50.22 & 44.80 & 60.98 & 91.04 & 69.52 & 57.38 & 12.77 \\[0.3ex]
  & Few-shot + CoT & 57.62 & 58.95 & 57.62 & 51.55 & 63.18 & 94.06 & 72.70 & 61.16 & 16.70 \\[0.3ex]
  & RAG & 55.02 & 63.51 & 55.02 & 52.02 & 58.01 & 94.85 & 68.84 & 56.23 & 11.86 \\[0.3ex]
  \rowcolor{mateGreen}
  & w/ {\small\datasetname{}} (SFT) & 73.61 & 82.17 & 73.61 & 75.20 & 55.99 & 92.26 & 66.85 & 55.26 & 12.29 \\[0.3ex]
  \rowcolor{mateGreenStrong}
  & w/ {\small\datasetname{}} (Full) & \textbf{88.98} & \textbf{91.11} & \textbf{88.98} & \textbf{88.63} & \textbf{66.70} & 91.45 & \textbf{74.24} & \textbf{63.03} & \textbf{18.40} \\[0.3ex]
  
  \hdashline[1pt/2pt]
  \multirow{6}{*}{Qwen3-14B}
  & Base & 42.31 & 66.35 & 42.31 & 38.67 & 60.66 & 93.62 & 70.26 & 57.86 & 12.50 \\[0.3ex]
  & Few-shot & 52.24 & 61.59 & 52.24 & 47.04 & 60.54 & 96.41 & 71.34 & 59.01 & 13.46 \\[0.3ex]
  & CoT & 50.94 & 59.34 & 50.94 & 46.92 & 61.96 & 91.66 & 69.88 & 57.57 & 12.37 \\[0.3ex]
  & Few-shot + CoT & 64.15 & 70.99 & 64.15 & 59.87 & 62.48 & 94.40 & 71.79 & 59.87 & 15.88 \\[0.3ex]
  & RAG & 68.48 & 77.69 & 68.48 & 68.40 & 61.23 & \textbf{97.15} & 72.18 & 59.89 & 13.78 \\[0.3ex]
  \rowcolor{mateGreen}
  & w/ {\small\datasetname{}} (SFT) & 74.33 & 75.27 & 74.33 & 70.39 & 63.94 & 93.37 & 72.99 & 61.31 & 17.11 \\[0.3ex]			
  \rowcolor{mateGreenStrong}
  & w/ {\small\datasetname{}} (Full) & \textbf{94.55} & \textbf{95.21} & \textbf{94.55} & \textbf{94.37} & \textbf{64.62} & 93.32 & \textbf{73.25} & \textbf{61.65} & \textbf{17.41} \\[0.3ex]
  \hline
  \end{tabular}%
  }
  
  \label{tab:mistake_classification}
  
  \end{table}

\subsection{The Dataset Provides the Right Supervision Signal}

We further explore which part of the framework is doing the real work. Table~\ref{tab:ablation_study} suggests a clear division of labor among the components. SFT accounts for most of the gain, indicating that task-specific supervision is the main source of mistake-detection capability. CoT annotations strengthen this competence, with the largest benefit appearing in violated principle classification, suggesting that explicit reasoning traces mainly help the model discriminate among subtle mistake categories. By contrast, GRPO brings smaller but consistent gains concentrated on sentence localization, implying that RL mainly calibrates output precision rather than teaching the task from scratch. The most striking drop comes from removing VGR: performance degrades sharply on both tasks, showing that data quality is not a cosmetic refinement but a necessary part of the supervision signal.

\begin{table*}[ht]
  \vspace{4pt}
  \caption{Ablation of SFT, CoT-augmented SFT, GRPO, Novice Growth Reward (NGR), and Validator-Guided Refinement (VGR), conducted on the Qwen3-8B model. ``--'' indicates NGR is not applicable without GRPO.}
  \centering
  
  \resizebox{\textwidth}{!}{%
  \renewcommand{\arraystretch}{1.1}
  \begin{tabular}{l c c c c c|c c c c|c c c c c}
  \hline
  \multirow{2}{*}{\textbf{Setting}} &
  \multirow{2}{*}{\textbf{VGR}} &
  \multirow{2}{*}{\textbf{SFT}} &
  \multirow{2}{*}{\textbf{CoT}} &
  \multirow{2}{*}{\textbf{GRPO}} &
  \multirow{2}{*}{\textbf{NGR}} &
  \multicolumn{4}{c|}{\textbf{Task 1: Violated Principle Classification}} &
  \multicolumn{5}{c}{\textbf{Task 2: Ethical Violation Location}} \\[0.5ex]
  \cline{7-10}\cline{11-15}
  & & & & & &
  {Accuracy} & {Precision} & {Recall} & {F1-Score} &
  {Precision} & {Recall} & {F1-Score} & {Jaccard} & {EMR} \\[0.5ex]
  \hline
  
  Baseline
  &  &  &  &  &
  & 50.22 & 57.03 & 50.22 & 44.80
  & 60.98 & 91.04 & 69.52 & 57.38 & 12.77 \\[0.3ex]
  
  \textbf{Full}
  & \cmark & \cmark & \cmark & \cmark & \cmark
  & \textbf{88.98} & \textbf{91.11} & \textbf{88.98} & \textbf{88.63}
  & \textbf{66.70} & 91.45 & \textbf{74.24} & \textbf{63.03} & \textbf{18.40} \\[0.3ex]
  
  w/o NGR
  & \cmark & \cmark & \cmark & \cmark & \xmark
  & 87.71 & 89.97 & 87.71 & 87.21
  & 64.02 & 94.09 & 73.17 & 61.34 & 15.92 \\[0.3ex]
  
  w/o SFT
  & \cmark & \xmark & \xmark & \cmark & \cmark
  & 52.42 & 59.74 & 52.42 & 47.28
  & 60.96 & 90.23 & 69.19 & 57.29 & 12.78 \\[0.3ex]
  
  w/o CoT
  & \cmark & \cmark & \xmark & \cmark & \cmark
  & 70.25 & 79.77 & 70.25 & 68.92
  & 59.51 & \textbf{96.83} & 71.17 & 58.75 & 12.24 \\[0.3ex]
  
  w/o GRPO
  & \cmark & \cmark & \cmark & \xmark & --
  & 73.61 & 82.17 & 73.61 & 75.20
  & 55.99 & 92.26 & 66.85 & 55.26 & 12.29 \\[0.3ex]
  
  w/o VGR
  & \xmark & \cmark & \cmark & \cmark & \cmark
  & 54.65 & 79.00 & 54.65 & 59.62
  & 41.71 & 61.31 & 46.84 & 39.29 & 11.12 \\[0.3ex]
  
  \hline
  \end{tabular}%
  }
  \label{tab:ablation_study}
  \vspace{-3mm}
\end{table*}

\subsection{The Supervisor Improves Human Training Experience}

\begin{figure}[!t]
  \centering
  \begin{minipage}{0.5\columnwidth}
      \centering
      \includegraphics[width=\linewidth]{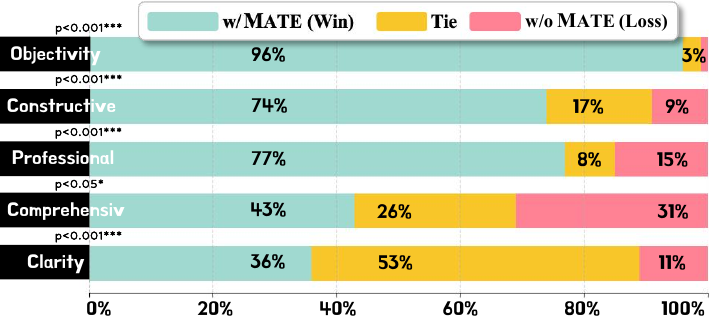}
  \caption{LLM evaluation results comparing the win, loss, and tie rates of critiques generated by Qwen3-8B fine-tuned on \datasetname{} (Win) against the base model (Loss). Results are statistically significant ($p < 0.05$, Bootstrap).}
      \label{fig:llm_judge_results}
  \end{minipage}
  \hfill  
  \begin{minipage}{0.47\columnwidth}
      \centering
      \includegraphics[width=\linewidth]{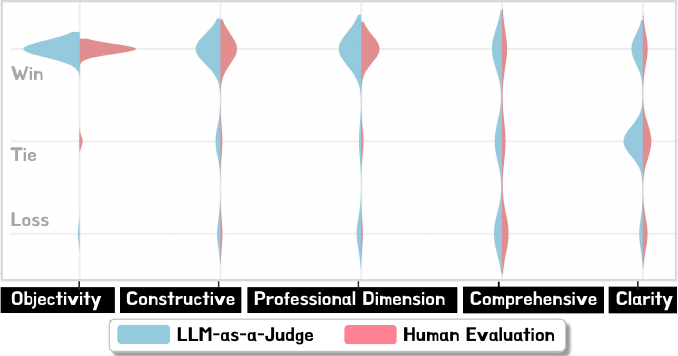}
      \caption{High agreement between LLM and human evaluation in supervisory feedback assessment across evaluated criteria.}
      \label{fig:xiaotiqin}
  \end{minipage}
  \vspace{-5mm}
  \end{figure}

\textbf{Feedback quality evaluation.}
We first examine whether the generated feedback itself is judged as better supervision. Following the LLM-as-a-Judge paradigm~\cite{lee2024cactus}, we use Skywork-Reward-Llama-3.1-8B~\cite{liu2024skywork} for pairwise comparisons on 300 test samples across five criteria: Objectivity \& Fairness, Constructiveness \& Actionability, Professional Depth, Comprehensiveness, and Clarity \& Structure. As shown in Figure~\ref{fig:llm_judge_results}, fine-tuning with \datasetname{} significantly improves critique quality across these professional dimensions, with especially clear gains in objectivity, professional depth, and constructiveness.

To validate the reliability of these automatic preference results, we further compare them with manual expert evaluation. Two psychology experts re-conducted pairwise scoring on the same samples, and Figure~\ref{fig:xiaotiqin} shows high consistency between the automatic and manual evaluations. This suggests that the improvements observed above are not artifacts of the reward model, but remain visible under human professional judgment.

\begin{wrapfigure}{r}{0.45\textwidth}
  \vspace{-16pt}
  \centering
  \includegraphics[width=\linewidth]{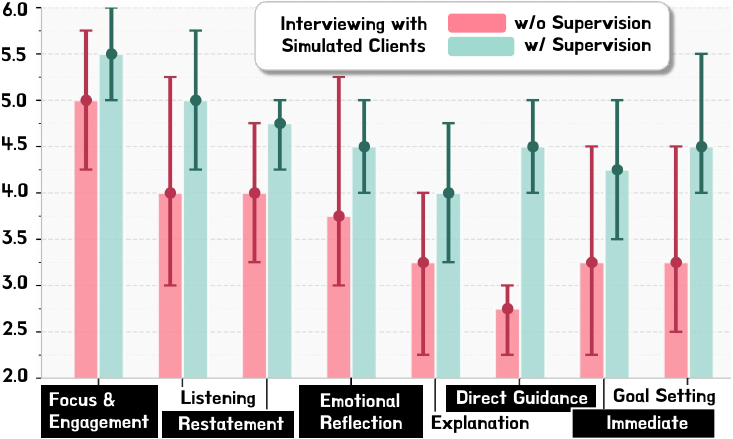}
  \caption{Self-Efficacy of Novice Counselors Before and After Supervised Feedback.}
  \label{fig:self-eff}
  \vspace{-10pt}
\end{wrapfigure}

\textbf{Trainee experience evaluation.}
We further test whether automated supervision improves the trainee's own experience by strengthening novice self-efficacy, a key indicator of perceived counseling competence~\cite{larson1998review, wang2025exploring, tang2025enhancing}. We built an online platform where novice counselling-psychology students conduct text-based sessions with an LLM-driven virtual client and then receive feedback from the fine-tuned supervisor. Figure~\ref{fig:self-eff} visualizes the resulting changes in novice self-efficacy.

Six counselling-psychology graduate students (3 women, 3 men; age $M_{\text{age}} = 23$; all with fewer than twenty hours experience) participated in the study. All participants held undergraduate degrees in psychology or related fields. Three students had previous volunteer experience in mental health hotlines, and two reported having sought personal psychological counselling in the past. Each student completed the CASES-R \cite{hahn2021assessment, hunsmann2024basic} twice: once before reading feedback and again after a five-minute reflection period. We focused on eight items from the exploration--insight and action subscales, each rated from 1 (no confidence) to 6 (complete confidence), yielding 48 paired observations.
Scores increased significantly on all eight skills after feedback
(10,000 sample bootstrap, 95\% CIs).
Gains were largest for procedural skills of \emph{Direct Guidance} and
\emph{Goal Setting}, whereas \emph{Focus}, already near the
ceiling, improved only slightly. Interview data indicated that initial
overconfidence in \emph{Reflection} produced a ceiling effect that
dampened its measurable improvement \citep{staus2021addressing,sabella2024pervasive}.

Overall, the critical-feedback framework substantially enhanced novices' confidence across a broad range of generic and task-specific counselling skills, offering encouraging evidence for the integration of automated supervision into counselor training. Together, these gains suggest that the supervisor functions as a practical scaffold for novice counselors: after receiving targeted feedback, trainees report stronger confidence not only in general counseling skills but also in task-specific intervention planning. This provides human-facing evidence that the proposed supervision framework can help build a learning ladder from feedback to perceived clinical readiness.

\section{Discussion \& Conclusion}
\label{sec:discussion}

This work reframes mental-health AI from a patient-facing counselor to an educational supervisor, proposing a \emph{Do No Harm} supervision paradigm grounded in the zone of proximal development and scaffolding theories: the supervisor helps novice counselors recognize ethically risky responses, understand why they are harmful, and move toward safer practice. To instantiate this paradigm, we construct \datasetname{}, a 9,915-instance human-in-the-loop dialogue-feedback dataset built from expert-distilled ethical principles, novice-error simulation, Validator-Guided Refinement, and expert review, covering ethical violation location, violated principle classification, and explanatory feedback. Methodologically, we train supervisors through SFT followed by GRPO with a Novice Growth Reward, optimizing feedback not merely for correctness but for whether it helps a weaker novice improve after reading the explanation. Experiments show consistent gains across supervision and training outcomes: Qwen3-14B reaches 94.37\% F1 on violated principle classification, a 144.1\% relative improvement over its base model and 29.1\% over GPT-4o with RAG, while Qwen3-8B achieves the best problematic-sentence localization F1 of 74.24\%, improving 38.6\% over its base model; downstream counselor evaluation further shows that our feedback improves novice performance across all six clinical metrics, including a 29.5\% relative gain on WAI over the unguided novice. Human-study results additionally indicate significant self-efficacy improvements across all eight assessed counseling skills after AI supervisory feedback. Together, these findings suggest that ethics-centered AI supervision can serve as a scalable scaffold for counselor training, shifting LLMs toward a safer and more constructive role: not replacing clinicians, but helping novices internalize the professional constraints needed to do no harm.

\bibliographystyle{plainnat}  
\bibliography{ref}


\end{document}